%% file: acl_latex.tex
\pdfoutput=1

\documentclass[11pt]{article}

\usepackage{acl}

\usepackage{times}
\usepackage{latexsym}

\usepackage[T1]{fontenc}

\usepackage[utf8]{inputenc}

\usepackage{microtype}

\usepackage{graphicx}
\usepackage{amsmath}
\usepackage{pifont}%
\newcommand{\cmark}{\ding{51}}%
\newcommand{\xmark}{\ding{55}}%

\usepackage{booktabs}
\usepackage{xcolor}
\usepackage{float}
\usepackage{makecell}
\usepackage{multirow}
\usepackage[multiple]{footmisc}
\usepackage[normalem]{ulem}
\usepackage{xurl}
\usepackage{subcaption}
\usepackage{mwe}

\usepackage{enumitem,amssymb}
\newlist{todolist}{itemize}{2}
\setlist[todolist]{label=$\square$}
\usepackage{pifont}

\def\flores{FLORES\xspace}
\def\muse{MUSE\xspace}
\def\mlqa{MLQA\xspace}
\def\xquad{XQuAD\xspace}
\def\sib{SIB\xspace}

\def\mt{MT\xspace}
\def\tlm{TLM\xspace}
\def\xss{XSS\xspace}

\definecolor{posred}{RGB}{201, 102, 87}
\definecolor{neggreen}{RGB}{32, 178, 164}

\title{A Recipe of Parallel Corpora Exploitation for Multilingual Large Language Models}

\author{Peiqin Lin$^{1,2}$, André F. T. Martins$^{3,4,5}$, Hinrich Schütze$^{1,2}$ \\
        $^1$Center for Information and Language Processing, LMU Munich \\
        $^2$Munich Center for Machine Learning \\
        $^3$Instituto Superior Técnico, Universidade de Lisboa (Lisbon ELLIS Unit) \\
        $^4$Instituto de Telecomunicações \quad
        $^5$Unbabel \\
        \texttt{linpq@cis.lmu.de}
}

\newcounter{notecounter}
\newcommand{\enotesoff}{\long\gdef\enote##1##2{}}

\enotesoff

\begin{document}
\maketitle
\begin{abstract}
Recent studies have highlighted the potential of exploiting parallel corpora to enhance multilingual large language models in both bilingual tasks, e.g., machine translation, and general-purpose tasks, e.g., text classification.
Building upon these findings, our comprehensive study aims to identify the most effective strategies %
for leveraging parallel corpora.
We investigate the impact of parallel corpus quality and quantity, training objectives, and model size on the performance of multilingual large language models enhanced with parallel corpora across diverse languages and tasks.
Our analysis reveals several key insights:
(i) filtering noisy translations is essential for exploiting parallel corpora, while language identification and short sentence filtering have little effect;
(ii) even a corpus with just 10K parallel sentences can yield results comparable to those obtained from larger datasets;
(iii) employing only the machine translation objective yields the best results among various training objectives and their combinations;
(iv) larger multilingual language models benefit more from parallel corpora than smaller models.
Our study offers valuable insights into the optimal utilization of parallel corpora to enhance multilingual large language models, extending the generalizability of previous findings from limited languages and tasks to a broader range of scenarios.
\end{abstract}

\enote{hs}{abstract/intro can probably be made more
exciting. i could give that a try tomorrow if you want ot}

\section{Introduction}

\begin{figure}[ht]
  \centering
  \resizebox {\columnwidth} {!} {
    \includegraphics[clip]{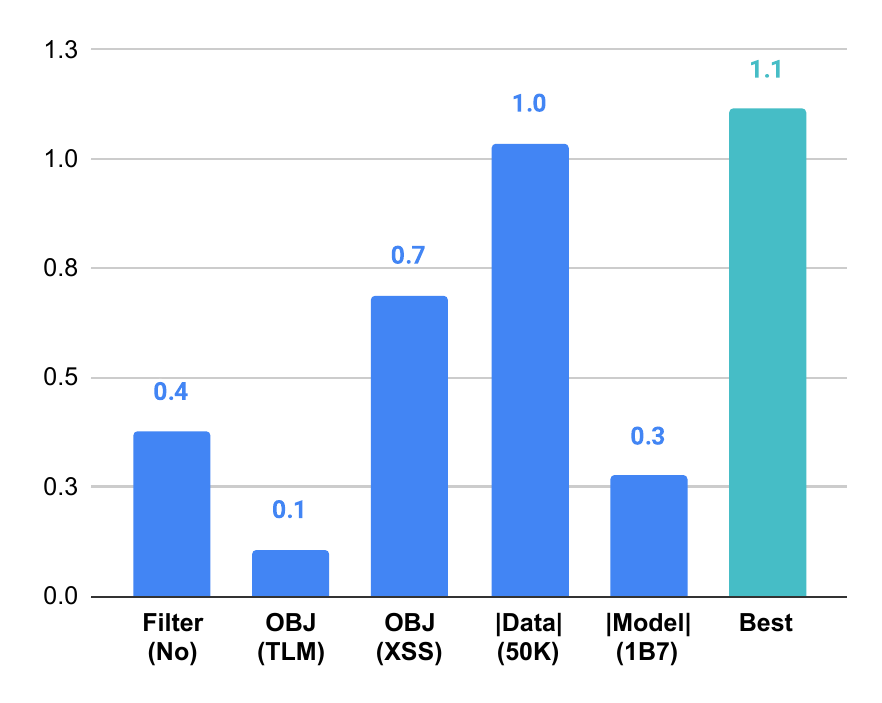}
  }
  \caption{
  Average performance improvements (y-axis) achieved by mLLMs
  enhanced with parallel corpora compared to their base
  models. \textbf{Best: Instruction tuning of BLOOM-7B1 with
  the machine translation objective (\mt) using 10K
  high-quality (i.e., filtered) parallel sentences yields the best
  results}. Main variations explored include: \textbf{Filter
  (No)} (using the original data); \textbf{OBJ (\tlm)}
  (translation language modeling objective);
  \textbf{OBJ (\xss)} (cross-lingual semantic similarity
  objective);
\textbf{|Data| (50K)}
(a larger 50K-sentence dataset); \textbf{|Model| (1B7)} (BLOOM-1B7 model).}
  \label{fig:overview}
\end{figure}

Recent multilingual large language models (mLLMs), represented by BLOOM \citep{DBLP:journals/corr/abs-2211-05100}, MaLA500 \citep{DBLP:journals/corr/abs-2401-13303}, and Aya \citep{DBLP:journals/corr/abs-2402-07827}, have shown impressive capacity on diverse tasks across languages. Parallel corpora have emerged as crucial resources for enhancing mLLMs, both for specific tasks, e.g., machine translation \citep{DBLP:journals/corr/abs-2309-11674,DBLP:journals/corr/abs-2402-17733}, and for general-purpose tasks \citep{DBLP:journals/corr/abs-2305-13627,DBLP:journals/corr/abs-2308-04948,DBLP:journals/corr/abs-2311-08089}.

However, existing studies often lack in comprehensive exploration of methodologies for harnessing parallel corpora. The quality and quantity of parallel corpora remain inadequately explored, inhibiting the full potential of such resources. Moreover, the influence of different training objectives and mLLM sizes across diverse languages and tasks remains under-investigated. This limitation impedes the generalization of parallel corpus exploitation methods across varied linguistic landscapes and task domains. Therefore, this paper aims to address these gaps by presenting a comprehensive recipe for exploiting parallel corpora for mLLMs. We focus on four key factors, with some main results shown in Figure \ref{fig:overview}.

\textbf{Quality:} %
We explore three dimensions of parallel corpus quality: translation accuracy, sentence length, and language identification. Our results show that \emph{translation quality}
is vital for exploiting parallel corpora, while sentence length filtering and language identification have minimal impact.

\textbf{Quantity:} Acquiring large amounts of high-quality parallel corpora is challenging, especially for relatively low-resource languages. Our study examines the minimum corpus size necessary to achieve performance improvements across diverse tasks. We find that even a corpus of just \emph{10K sentences} can yield results comparable to those obtained from much larger datasets.

\textbf{Objective:} Previous studies \citep{DBLP:journals/corr/abs-2305-13627} have investigated the effectiveness of different training objectives and their combinations on classification tasks of Indonesian local languages, using smaller-sized mLLMs up to 1B7 parameters. We extend this investigation by examining the impact of various training objectives and their combinations on larger mLLMs across a range of languages and tasks. Our experiments demonstrate that employing \emph{the machine translation objective} produces the most promising results.

\textbf{Model Size:} The size of mLLMs can greatly impact their ability to comprehend instructions derived from parallel corpora. Our findings indicate that \emph{larger mLLMs} exhibit superior comprehension and cross-task transferability compared to their smaller counterparts. Consequently, they achieve more substantial improvements across a broader spectrum of tasks.

In light of the critical role parallel corpora play in mLLMs, our study provides a comprehensive recipe for effectively exploiting parallel corpora. We have identified four primary factors: quality (\S\ref{sec:quality}), quantity (\S\ref{sec:quantity}), objective (\S\ref{sec:objective}), and model size (\S\ref{sec:model_size}). 
Our detailed analysis of these factors reveals their great impact on mLLM performance across diverse languages and tasks. By delving into these aspects, we offer actionable insights that can inform the development and optimization of strategies for parallel corpus exploitation, ultimately contributing to the advancement of mLLMs in both bilingual and general-purpose tasks.

\section{Related Work}

\subsection{Parallel Data for Multilingual Language Models}

Over the years, multilingual language models have evolved from earlier, smaller models, such as XLM \citep{DBLP:conf/nips/ConneauL19}, XLM-R \citep{DBLP:conf/acl/ConneauKGCWGGOZ20}, and Glot500 \citep{DBLP:conf/acl/ImaniLKSSKMSMYS23}, to more recent, larger models, including BLOOM \citep{DBLP:journals/corr/abs-2211-05100}, MaLA500 \citep{DBLP:journals/corr/abs-2401-13303}, and Aya \citep{DBLP:journals/corr/abs-2402-07827}. These models consistently demonstrate strong performance across various downstream tasks \citep{DBLP:journals/corr/abs-2303-12528,DBLP:journals/corr/abs-2405-05116}.

Parallel corpora have played a pivotal role in both the analysis \citep{DBLP:conf/coling/PiquerasS22,DBLP:conf/eacl/LinHZMS24} and enhancement \citep{DBLP:conf/nips/ConneauL19,DBLP:journals/corr/abs-2012-15674,DBLP:conf/aaai/YangMZWL020,DBLP:conf/emnlp/HuangLDGSJZ19,DBLP:conf/naacl/ChiDWYSWSMHZ21,DBLP:conf/iclr/WeiW0XYL21,DBLP:conf/naacl/HuJFSN21,DBLP:conf/acl/Chi0ZHMHW20,DBLP:conf/naacl/ReidA22,DBLP:journals/corr/abs-2311-08849} of small multilingual language models.

In the era of mLLMs, parallel corpora are constructed as instruction data and used to enhance mLLMs through supervised fine-tuning \citep{DBLP:journals/corr/abs-2305-13627,DBLP:journals/corr/abs-2308-04948,DBLP:journals/corr/abs-2311-08089}. Specifically, \citet{DBLP:journals/corr/abs-2305-13627} propose three methods of incorporating parallel corpora as instruction tuning data: Machine Translation (\mt), Translation Language Modeling (\tlm), and Cross-Lingual Semantic Similarity (\xss) (see \S\ref{sec:training}). However, their evaluation is limited to small models with up to 1.7 billion parameters and focuses solely on classification tasks within Indonesian local languages. Both \citet{DBLP:journals/corr/abs-2308-04948} and \citet{DBLP:journals/corr/abs-2311-08089} propose using machine-translation-style instruction data to improve mLLMs but do not explore different training objectives. While these studies yield promising results, their scope is limited. Firstly, they do not explore critical factors such as the quality and quantity of parallel corpora, considering the high cost of collecting high-quality and massive parallel corpora, especially for relatively low-resource languages. Secondly, their investigations do not encompass an in-depth analysis of training objectives and mLLMs with varied model sizes across diverse languages and tasks.

\enote{hs}{for XSS, what are the possible labels? are they
numerical? just yes, no, unsure?}

\enote{pl}{yes/no. I added a notes in the paper: see 3.3}

\subsection{Key Elements for Language Modeling}

Previous research has extensively examined critical factors essential for the pretraining and enhancement of language models.

\textbf{Quality:} \citet{DBLP:journals/tacl/KreutzerCWWEUTS22}
conducted manual audits of prevalent monolingual and parallel corpora, revealing significant portions of low-quality data, particularly in corpora for relatively low-resource languages. Follow-up studies have investigated the impact of data quality on model performance. \citet{DBLP:conf/emnlp/ArtetxeAAPS22} observed
that similar results on downstream tasks can be achieved
regardless of the degree of quality of the corpus used for pretraining, while other studies found that the quality of parallel corpora matters for machine translation \citep{DBLP:journals/corr/abs-2402-07446} and general-purpose tasks \citep{DBLP:journals/corr/abs-2212-10173}.

\textbf{Quantity:} Recent works \citep{DBLP:journals/corr/abs-2305-09246,DBLP:journals/corr/abs-2305-11206,DBLP:journals/corr/abs-2306-05539} have focused on the impact of fine-tuning with small amounts of high-quality instruction data, such as one or a few thousand instances, showing promising performance gains in evaluation tasks. \citet{DBLP:journals/corr/abs-2309-11674} demonstrate that as few as 10K high-quality parallel sentences can significantly enhance machine translation performance.

\textbf{Objective:} Different training objectives based on parallel corpora for enhancing mLLMs can be viewed as distinct instructions. \citet{DBLP:journals/corr/abs-2306-04751} explore the impact of various types of instruction tuning data and find that their combination can be optimal in certain scenarios.

\textbf{Model Size:} Recent studies indicate that scaling up language models enhances their capability to excel in diverse and complex reasoning tasks \citep{DBLP:journals/tmlr/WeiTBRZBYBZMCHVLDF22,DBLP:journals/corr/abs-2303-03846,DBLP:journals/corr/abs-2309-01809}. Follow-up studies %
\citep{DBLP:journals/corr/abs-2303-03846} further illustrate distinct behavioral differences between larger and smaller models.

However, these factors have not yet been comprehensively explored in the context of leveraging parallel corpora to enhance mLLMs across diverse languages and tasks.

\section{Setup}
\label{sec:setup}

\subsection{Language}

We use three criteria for language selection. Firstly,
we select
languages well covered by mLLMs as our goal is to assess how parallel data enhances performance after pre-training an mLLM with monolingual data. Secondly, the selected languages should be also covered by several different evaluation
benchmarks, allowing for robust evaluation across diverse
downstream tasks. Lastly, we select typologically
diverse languages, enabling our investigation to generalize to a wide
range of relatively low-resource languages. Thus, we select five languages: Arabic (ar),
Spanish (es), Hindi (hi), Vietnamese (vi) and Chinese (zh).

\subsection{Data}

We utilize the OPUS100 dataset \citep{DBLP:conf/acl/ZhangWTS20}, an English-centric multilingual corpus, to gather parallel sentences between English (en) and each target language. The quality of OPUS100 is assessed across three dimensions:

\paragraph{Translation Quality} Manual quality assessment of the vast amount of parallel corpora is impractical. Instead, we employ COMETKIWI \citep{DBLP:conf/wmt/ReiTGZFMSGACLM22}\footnote{\url{https://huggingface.co/Unbabel/wmt23-cometkiwi-da-xxl}}, a tool for estimating the quality of machine translation outputs across multiple languages. We set a COMETKIWI score threshold $\tau_c$, retaining parallel corpora with scores not lower than $\tau_c$.

\paragraph{Sentence Length} Given the variation in character length across languages, we avoid using it as a metric for consistency. Instead, we measure sentence length by the number of tokens, as determined by the tokenizer of our chosen mLLM, BLOOM-7B1. We establish a length threshold $\tau_l$, retaining parallel corpora where both source and target sentences contain no fewer than $\tau_l$ tokens.

\paragraph{Language Identification} To identify sentences potentially not in the correct language, we employ GlotLID \citep{DBLP:conf/emnlp/KargaranIYS23}, an open-source language identification model. This language identification filter is applied to both the source and target sentences.

\begin{table*}[ht]
    \centering
    \resizebox{\textwidth}{!}{
    \begin{tabular}{c|p{\linewidth}}
        \toprule
        Objective & Template \\
        \midrule
        \mt & \texttt{Translate the following text from [SOURCE\_LANG] to [TARGET\_LANG].\textbackslash nText: [SOURCE\_TEXT]\textbackslash nTranslation: [TARGET\_TEXT]} \\
        \midrule
        \tlm & \texttt{[INPUT\_TEXT]. Denoise the previous [TARGET\_LANG] text to its equivalent sentence in [SOURCE\_LANG]: [SOURCE\_TEXT]\textbackslash n[TARGET\_TEXT]} \\
        \midrule
        \xss & \texttt{[SOURCE\_LANG] sentence: [SOURCE\_TEXT]\textbackslash n[TARGET\_LANG] sentence: [TARGET\_TEXT]\textbackslash nDo the two sentences have the same meaning? [LABEL]} \\
        \bottomrule
    \end{tabular}
    }
    \caption{Templates of \mt, \tlm, and \xss for instruction data construction based on parallel corpora.}
    \label{tab:template}
\end{table*}

\subsection{Training}
\label{sec:training}

We select the BLOOM
series \citep{DBLP:journals/corr/abs-2211-05100} for our
investigation due to its offering of different sizes of
mLLMs which well cover the five target languages under
consideration.\footnote{Additional experiments with XGLM, as presented in \S\ref{sec:xglm}, yield conclusions consistent with those of BLOOM.} The pretraining data size for the five target languages ranges from 23GB (Hindi) to 452GB (English). We explore BLOOM models of various sizes,
including 7B1, 3B, and 1B7. Due to limited computational
resources, we use LoRA \citep{DBLP:conf/iclr/HuSWALWWC22}, which is known for
its competitive performance compared to full-parameter
training \citep{DBLP:conf/emnlp/AlvesGAPRSCM23}, for instruction tuning of BLOOM. We configure the learning rate to $1\times 10^{-4}$, weight
decay to $0.1$, and set the rank of LoRA to $16$ based on preliminary experiment in \S\ref{sec:lora}. The
maximum sequence length for both source and target sentences
is set to $128$. To maintain consistency across experiments
with different quantities of parallel corpora, we ensure a
uniform training budget of 50K parallel
sentences. Specifically, we calculate the number of epochs
as 50K divided by the number of sentences considered from the OPUS100 dataset. The batch size is $128$, and we save the checkpoints every 20 steps.

Following \citet{DBLP:journals/corr/abs-2305-13627}, we construct the data for instruction tuning based on the parallel corpora by three distinct patterns: Machine Translation (\mt), Translation Language Modeling (\tlm), and Cross-Lingual Semantic Similarity (\xss). Table~\ref{tab:template} presents the templates for these three objectives. Here, \texttt{[SOURCE\_LANG]} and \texttt{[TARGET\_LANG]} represent the language names of the source and target languages, respectively. In our study, we consider both English-to-target-language and target-language-to-English directions, where \texttt{[SOURCE\_LANG]} represents English or \texttt{[TARGET\_LANG]} represents English. For \mt, \texttt{[SOURCE\_TEXT]} and \texttt{[TARGET\_TEXT]} refer to the parallel sentences in the source and target languages, respectively. For TLM, a portion of tokens in \texttt{[TARGET\_TEXT]} are masked to generate \texttt{[INPUT\_TEXT]}. For XSS, our objective is to predict whether parallel sentences \texttt{[SOURCE\_TEXT]} and \texttt{[TARGET\_TEXT]} are semantically similar, with \texttt{[LABEL]} being ``Yes'' or ``No''. Specifically, we utilize the parallel corpora as positive examples and introduce perturbations to \texttt{[TARGET\_TEXT]} to construct negative examples. We consider applying the objectives both individually and in combination. We tune the model with the objective of causal language modeling without loss mask.

\subsection{Evaluation}

We conduct evaluation across five diverse benchmarks: \flores \citep{DBLP:journals/corr/abs-2207-04672}, \muse \citep{DBLP:conf/iclr/LampleCRDJ18}, MLQA \citep{DBLP:conf/acl/LewisORRS20}, XQUAD \citep{DBLP:conf/acl/ArtetxeRY20}, and \sib \citep{DBLP:journals/corr/abs-2309-07445}. A comprehensive overview of these benchmarks is available in Table~\ref{tab:benchmark}. Our evaluation spans both classification tasks (\sib) and generation tasks (\flores, \muse, \mlqa, and \xquad), covering a spectrum of cross-language (\flores, \muse, and \mlqa) and in-language tasks (\xquad and \sib).

For translation tasks within \flores and \muse, we explore bidirectional translation: from English to other languages (en-xx) and from other languages to English (xx-en). Additionally, for \mlqa, we evaluate scenarios where questions are in English and the passages and answers are in other languages (en-xx), as well as situations where questions are in other languages and the passages and answers are in English (xx-en).

To provide a thorough understanding of our evaluation procedures, we offer detailed prompts for each task in \S\ref{sec:prompt}. In all experiments, we employ a 2-shot in-context learning approach, where the model is given two examples appended to the input query to aid in making predictions.

\begin{table}[]
    \centering
    \resizebox{\columnwidth}{!}{
    \begin{tabular}{c|ccccc}
        \toprule
        Dataset & Task & |Data| & Metric & I/C & C/G \\
        \midrule
        \flores & Machine Translation & 1012 & COMETKIWI & C & G \\
        \muse & Word Translation & 1500 & F1 & C & G \\
        \mlqa & Question Answering & 4918 - 5495 & F1 & C & G \\
        \xquad & Question Answering & 1190 & F1 & I & G \\
        \sib & Text Classification & 204 & Acc & I & C \\
        \bottomrule
    \end{tabular}
    }
    \caption{Details of evaluation benchmarks. |Data|: Number of samples for evaluation. I/C: In-language/Cross-language. C/G: Classification/Generation.}
    \label{tab:benchmark}
\end{table}

\begin{figure}[]
  \centering
  \resizebox {\columnwidth} {!} {
    \includegraphics[clip]{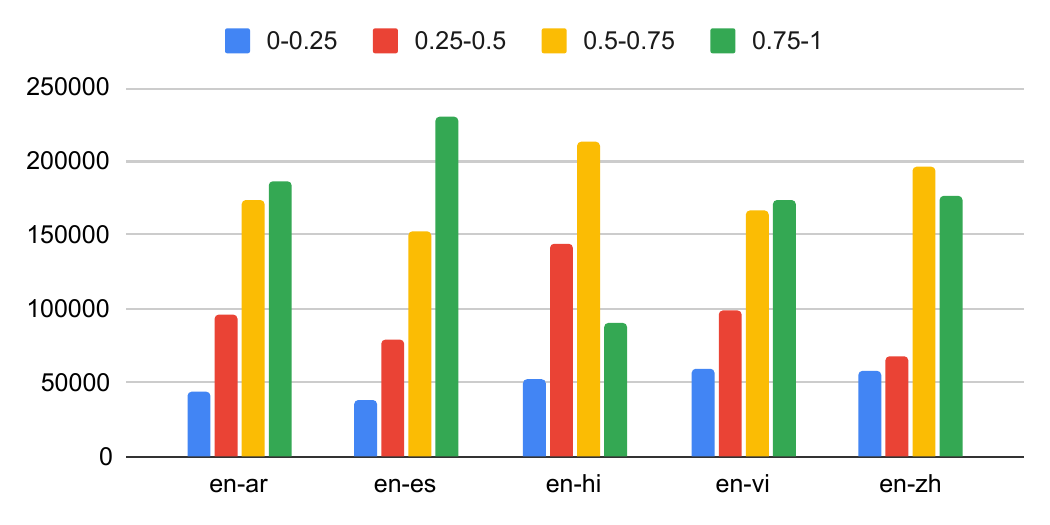}
  }
  \caption{Translation quality measured by COMETWIKI of
  500K parallel sentences from OPUS100 for our five  language pairs. The COMETWIKI scores are segmented into four ranges: 0-0.25, 0.25-0.5, 0.5-0.75, and 0.75-1. Higher scores represent better translation quality.}
  \label{fig:comet}
\end{figure}

\begin{figure}[ht]
  \centering
  \resizebox {\columnwidth} {!} {
    \includegraphics[clip]{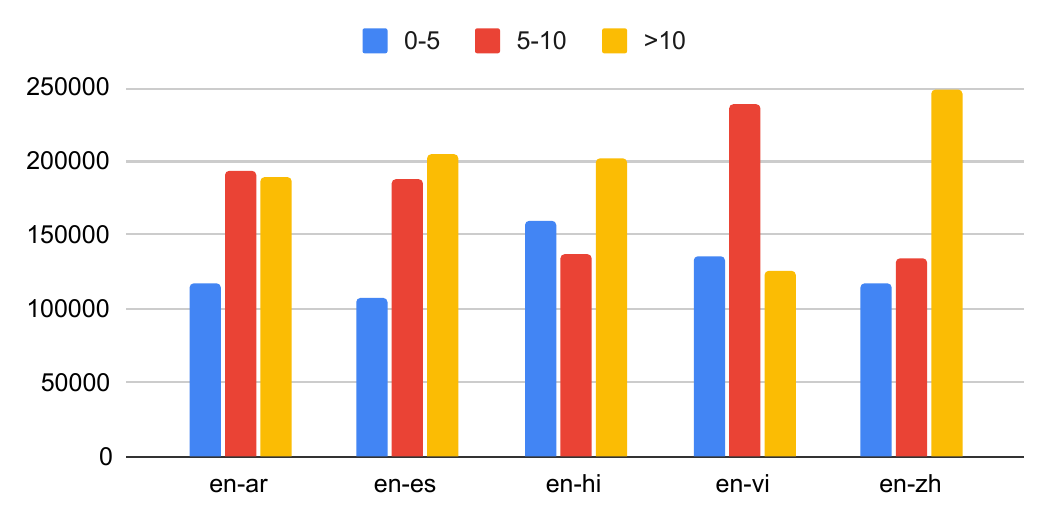}
  }
  \caption{Sentence length of 500K parallel sentences
  from OPUS100 for our five language pairs. The three
  categories are 0-5, 5-10, greater than 10 tokens.}
  \label{fig:len}
\end{figure}

\begin{figure}[ht]
  \centering
  \resizebox {0.95\columnwidth} {!} {
    \includegraphics[clip]{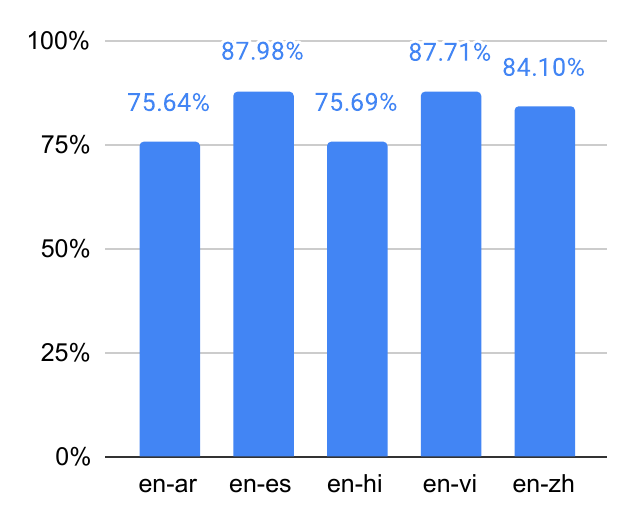}
  }
  \caption{Percentage of sentences retained after language
  identification filtering of 500K parallel sentences
  from OPUS100 for our five language pairs.}
  \label{fig:lid}
\end{figure}

\section{Quality}
\label{sec:quality}

\subsection{Quality of OPUS100}

We measure the quality of 500K parallel sentences from
OPUS100 for our five language pairs using three key metrics:
translation quality, sentence length, and language
identification accuracy, as illustrated in
Figures~\ref{fig:comet}--\ref{fig:lid}.

\textbf{A considerable portion of OPUS100 is of
low-quality}. All quality measures indicate that a large
portion of OPUS100 contains low-quality data. Approximately 10\% of the data has COMETKIWI scores below 0.25, indicating very poor translation quality. Additionally, between 10\% to 30\% of the data falls within the 0.25-0.5 score range, which is still considered sub-optimal. Regarding sentence length, we find that over 20\% of the OPUS100 data consists of very short sentences, with a length of no more than five tokens. For language identification, 13\%
to 25\% of the data is removed due to incorrect language
identification results in one of the two parallel sentences.

\textbf{Relatively low-resource languages suffer more from low-quality issues}. For relatively low-resource languages like Hindi, there are fewer high-quality parallel sentences compared to high-resource languages such as Spanish. Analysis of translation quality indicates that the English-Hindi pair has less than 20\% of parallel sentences with high COMETWIKI scores (0.75-1), whereas the English-Spanish pair has around 45\%. For sentence length, the English-Hindi pair contains 10\% more short sentences (0-5 tokens) compared to high-resource language pairs. Moreover, both the English-Arabic and English-Hindi pairs exhibit about 10\% more parallel sentences that may be in the wrong languages.

These comprehensive findings underscore the critical
importance of data quality when exploiting
parallel corpora for mLLM training.

\subsection{Effect of Quality}

\begin{table*}[ht]
\centering
\resizebox{0.95\textwidth}{!}{
\begin{tabular}{c|ccc|ccccccccc}
\toprule
\multirow{2}{*}{\textsc{ID}} & \multicolumn{3}{c|}{\multirow{2}{*}{\textsc{Model}}} & \multicolumn{2}{c}{\textsc{\flores}} & \multicolumn{2}{c}{\textsc{\muse}} & \multicolumn{2}{c}{\textsc{\mlqa}} & \multirow{2}{*}{\textsc{\xquad}} & \multirow{2}{*}{\textsc{\sib}} & \multirow{2}{*}{\textsc{AVG}} \\
\cmidrule{5-10}
& & & & \textsc{en-xx} & \textsc{xx-en} & \textsc{en-xx} & \textsc{xx-en} & \textsc{en-xx} & \textsc{xx-en} & & & \\
\midrule
0 & \multicolumn{3}{c|}{BLOOM-7B1} & 69.1 & \textbf{72.4} & 43.1 & 53.7 & 36.4 & 42.7 & 47.2 & 58.1 & 52.8 \\
\midrule
& $\tau_c$ & $\tau_l$ & LID & & & & & & & & & \\
\midrule
1 & 0 & 0 & \cmark & 69.7 & 71.7 & 44.4 & 52.6 & 37.8 & 43.0 & 47.7 & 58.8 & 53.2 \\
2 & 0.5 & 0 & \cmark & 69.9 & 72.1 & 45.0 & 53.0 & \textbf{38.1} & 43.7 & 48.1 & \textbf{59.8} & 53.7 \\
3 & 0.75 & 5 & \cmark & 70.3 & 72.1 & \textbf{45.7} & 53.6 & \textbf{38.1} & 43.5 & 47.8 & 59.2 & 53.8 \\
4 & 0.75 & 0 & \xmark & \textbf{70.5} & 72.1 & 44.9 & 53.7 & 37.7 & \textbf{44.0} & \textbf{48.3} & 59.6 & 53.9 \\
5 & 0.75 & 0 & \cmark & 70.3 & 72.3 & 45.5 & \textbf{53.9} & 38.0 & 43.9 & \textbf{48.3} & 59.5 & \textbf{54.0} \\
\bottomrule
\end{tabular}
}
\caption{Performance (\%) of BLOOM-7B1
after
instruction tuning with 
the machine translation objective using 10K parallel corpora with various quality filtering strategies. Parameters include $\tau_c$ for COMETWIKI score threshold, $\tau_l$ for sentence length threshold, and LID indicating the adoption of language identification filtering.}
\label{tab:quality}
\end{table*}

Table~\ref{tab:quality} presents the performance of
BLOOM-7B1 after instruction tuning with the
machine translation objective, using 10K parallel corpora
with various quality filtering strategies.

\textbf{Parallel corpora containing noisy translations
still improve results.} Comparing the results of the
experiment with $\tau_c=0$ (ID 1) to the original model (ID
0), there's an average improvement of 0.4\% for all
tasks. The most notable improvements are observed in both bilingual tasks (en-xx) and in-language tasks. However, generating English for bilingual tasks yields degraded or marginally improved results. Experiment 1 exhibits 0.7\% and 1.1\% decrements in \flores and \muse respectively, with only a 0.3\% improvement in \mlqa.

\textbf{Filtering out noisy translations leads to notable improvements.} When $\tau_c=0.5$, the average performance rises from 53.2\% to 53.7\%. Further refinement to $\tau_c=0.75$ achieves an additional 0.3\% improvement. These improvements are consistently observed across all evaluated tasks. In the optimal setting (ID 5), there's a 1.2\% improvement compared to BLOOM-7B1 (ID 0). The improvements corroborate the reliability of COMETKIWI as a metric for filtering low-quality translations.

\textbf{Filtering short sentences yields slightly worse results than using unfiltered data.} The experiment with filtering short sentences (ID 3) achieves comparable or slightly worse results compared to that without filtering short sentences (ID 5). This suggests that short sentences, whether at the word or phrase level, may offer some benefits for sentence-level tasks.

\textbf{Using data with language identification filtering
results in only a 0.1\% improvement on average.} A
comparison of experimental outcomes with and without
language identification filtering (ID 4 and 5) reveals that
using data with language identification filtering yields
merely a 0.1\% improvement on average. The most notable
performance difference is observed in the \muse task, where
using data with language identification filtering leads to
improvements of 0.6\% (en-xx) and 0.2\% (xx-en). This marginal enhancement may be attributed to the presence of sentences in similar languages within OPUS100, which exhibit minor linguistic variations compared to the true language. These variations could potentially have a slight negative impact on word-level translations while having little impact on sentence-level tasks.

\section{Quantity}
\label{sec:quantity}

\subsection{Effect of Quantity Across Tasks}
\label{sec:quantity_task}

\begin{table*}[ht]
\centering
\resizebox{0.95\textwidth}{!}{
\begin{tabular}{c|ccccccccc}
\toprule
\multirow{2}{*}{\textsc{|Sent|}} & \multicolumn{2}{c}{\textsc{\flores}} & \multicolumn{2}{c}{\textsc{\muse}} & \multicolumn{2}{c}{\textsc{\mlqa}} & \multirow{2}{*}{\textsc{\xquad}} & \multirow{2}{*}{\textsc{\sib}} & \multirow{2}{*}{\textsc{AVG}} \\
& \textsc{en-xx} & \textsc{xx-en} & \textsc{en-xx} & \textsc{xx-en} & \textsc{en-xx} & \textsc{xx-en} & & & \\
\midrule
0 & 69.1 & \textbf{72.4} & 43.1 & 53.7 & 36.4 & 42.7 & 47.2 & 58.1 & 52.8 \\
1K & 70.0 & 72.2 & 45.3 & 53.6 & \textbf{38.2} & 43.6 & 47.9 & 59.2 & 53.8 \\
5K & 70.3 & 72.2 & 45.4 & 53.5 & \textbf{38.2} & 43.8 & 48.2 & \textbf{59.5} & 53.9 \\
10K & 70.3 & 72.3 & \textbf{45.5} & \textbf{53.9} & 38.0 & 43.9 & 48.3 & \textbf{59.5} & \textbf{54.0} \\
25K & 70.3 & 72.2 & 45.1 & 53.8 & 38.0 & \textbf{44.0} & \textbf{48.4} & \textbf{59.5} & 53.9 \\
50K & \textbf{70.4} & 72.2 & 45.1 & 53.8 & 38.1 & 43.7 & 48.3 & \textbf{59.5} & 53.9 \\
\bottomrule
\end{tabular}
}
\caption{Task performance (\%) of BLOOM-7B1 after
instruction tuning with the machine translation objective using varying amounts of parallel sentences, obtained with the best filtering strategy (ID 5) as shown in Table~\ref{tab:quality}. \textsc{|Sent|} indicates the number of parallel sentences used for instruction tuning, with \textsc{|Sent|=0} representing the original BLOOM-7B1 model.}
\label{tab:quantity_task}
\end{table*}

Based on Table~\ref{tab:quantity_task}, which shows the
performance of BLOOM-7B1 after
instruction tuning with 
the machine translation objective using different
amounts of parallel sentences, we can derive the following
key findings:

\textbf{Adding merely 1K parallel sentences helps.} Exploiting 1K parallel sentences for instruction tuning improves the overall average score by 1\%. This increase is observed across most tasks, with notable improvements in \flores (en-xx), \muse (en-xx), and \sib.

\textbf{Using 10K parallel sentences leads to the optimal performance.} The best overall performance is achieved with 10K parallel sentences, resulting in an average score of 54.0\%. This setting yields the highest scores in \muse and \sib. This aligns with the findings in \citet{DBLP:journals/corr/abs-2309-11674}.

\textbf{More data achieves comparable results.} Increasing the number of parallel sentences beyond 10K results in comparable performance. Specifically, using 25K or 50K parallel sentences yields average scores of 53.9\%, which are very close to the score obtained with 10K sentences.

The analysis suggests that instruction tuning with a moderate amount of parallel sentences (around 10K) yields the best overall improvement in performance for the BLOOM-7B1 model across various tasks.

\begin{table}[ht]
\centering
\begin{tabular}{r|ccccc}
\toprule
& ar & es & hi & vi & zh \\
\midrule
0 & 49.5 & 57.7 & 46.5 & 63.8 & 46.7 \\
1K & 50.8 & 58.1 & \textbf{47.7} & 64.3 & 47.8 \\
5K & 51.2 & 58.2 & 47.6 & 64.6 & \textbf{47.9} \\
10K & \textbf{51.3} & \textbf{58.4} & \textbf{47.7} & 64.6 & 47.8 \\
25K & 51.2 & 58.2 & \textbf{47.7} & \textbf{64.7} & 47.7 \\
50K & 51.2 & 58.2 & \textbf{47.7} & \textbf{64.7} & 47.6 \\
\bottomrule
\end{tabular}
\caption{Language performance (\%) of BLOOM-7B1
after
instruction tuning with 
the machine translation objective using varying amounts of parallel sentences, obtained with the best filtering strategy (ID 5) as shown in Table~\ref{tab:quality}. \textsc{|Sent|} indicates the number of parallel sentences used for instruction tuning, with \textsc{|Sent|=0} representing the original BLOOM-7B1 model.}
\label{tab:quantity_lang}
\end{table}

\begin{table*}[ht]
\centering
\resizebox{0.95\textwidth}{!}{
\begin{tabular}{c|ccccccccc}
\toprule
\multirow{2}{*}{\textsc{Model}} & \multicolumn{2}{c}{\textsc{\flores}} & \multicolumn{2}{c}{\textsc{\muse}} & \multicolumn{2}{c}{\textsc{\mlqa}} & \multirow{2}{*}{\textsc{\xquad}} & \multirow{2}{*}{\textsc{\sib}} & \multirow{2}{*}{\textsc{AVG}} \\
& \textsc{en-xx} & \textsc{xx-en} & \textsc{en-xx} & \textsc{xx-en} & \textsc{en-xx} & \textsc{xx-en} & & & \\
\midrule
BLOOM-7B1 & 69.1 & \textbf{72.4} & 43.1 & 53.7 & 36.4 & 42.7 & 47.2 & 58.1 & 52.8 \\
\midrule
\mt & \textbf{70.3} & 72.3 & \textbf{45.5} & \textbf{53.9} & \textbf{38.0} & 43.9 & \textbf{48.3} & 59.5 & \textbf{54.0} \\
\tlm & 67.2 & 72.2 & 44.3 & 53.0 & 36.3 & 44.4 & 47.6 & 58.7 & 53.0 \\
\xss & 69.4 & 72.2 & 43.7 & 53.5 & 37.0 & 44.2 & \textbf{48.3} & 60.0 & 53.5 \\
\mt+\tlm & 69.3 & 72.1 & 44.1 & 53.2 & 36.8 & 43.8 & 47.2 & 59.5 & 53.2 \\
\mt+\xss & 70.3 & 72.1 & 44.9 & 53.3 & 37.4 & 44.5 & 47.9 & \textbf{60.4} & 53.8 \\
\tlm+\xss & 67.7 & 72.2 & 43.0 & 52.5 & 34.9 & \textbf{45.6} & 48.2 & 60.0 & 53.0 \\
\mt+\tlm+\xss & 69.5 & 72.1 & 44.2 & 53.2 & 36.1 & 45.1 & 47.7 & 59.0 & 53.4 \\
\bottomrule
\end{tabular}
}
\caption{Performance (\%) of BLOOM-7B1
after
instruction tuning 
with different objectives and their combinations using 10K parallel corpora, obtained with the best filtering strategy (ID 5) as shown in Table~\ref{tab:quality}.}
\label{tab:objective}
\end{table*}

\subsection{Effect of Quantity Across Languages}
\label{sec:quantity_lang}

We delve deeper into the influence of parallel corpora quantity,
as depicted in Table~\ref{tab:quantity_lang}.

\textbf{Using 10K parallel sentences achieves optimal performance across most languages.} For the majority of languages, except Vietnamese (vi) and Chinese (zh), the highest performance is obtained with 10K parallel sentences. Even for Vietnamese and Chinese, leveraging 10K parallel sentences can yield comparable results. These observations align with the findings  in \S\ref{sec:quantity_task}.

\textbf{Different languages exhibit varying appetites for parallel corpora.} Across most languages, increasing the number of parallel sentences used for instruction tuning generally leads to incremental improvements in performance. However, for Hindi (hi) and Chinese (zh), transitioning from 1K to 10K parallel sentences does not yield improvement. This phenomenon may be attributed to BLOOM-7B1's limited proficiency in these languages compared to others, as reflected in the results of the original BLOOM-7B1 model (\textsc{|Sent|=0}).

\section{Objective}
\label{sec:objective}

We explore the effectiveness of different objectives and their combinations, with results
in Table~\ref{tab:objective}.

\textbf{BLOOM-7B1 performs well on English generation tasks.} The baseline BLOOM-7B1 model exhibits robust performance across a spectrum of evaluation tasks, notably excelling in English generation tasks such as \flores (xx-en) and \muse (xx-en). Further exploitation of parallel corpora fails to yield any discernible improvement.

\textbf{\mt emerges as the top performer.} The \mt objective consistently outperforms the baseline BLOOM-7B1 model, showcasing an average improvement of 1.2\%. Moreover, \mt achieves the highest performance in 5 out of 8 evaluated tasks.

\textbf{\tlm exhibits limited effectiveness.} While \tlm shows slight improvements on average (0.2\%), primarily driven by enhancements in tasks like \muse (en-xx), \mlqa (xx-en), \xquad, and \sib, it also leads to degradation in tasks including \flores and \muse (xx-en).

\textbf{\xss achieves strong performance for classification.} Using the \xss objective improves BLOOM-7B1 by 0.7\%, though it performs 0.5\% worse than \mt. The major decrease is observed in translation tasks, especially from English to other languages. However, \xss can still slightly improve translation tasks compared to BLOOM-7B1. Notably, \xss achieves 0.3\% better performance on \sib, highlighting its effectiveness for classification.

\textbf{Combining training objectives does not provide large benefits.} While combinations of different objectives can improve BLOOM-7B1 by 0.2\% to 1.0\%, none surpass the performance of using the \mt objective alone. The combination of \mt and \xss is the best among the combinations, slightly worse than \mt by 0.2\%, but better than all other objectives. Notably, \mt+\xss achieves the best results on \sib, and \tlm+\xss yields the best results on \mlqa (xx-en). These observations indicate that no single objective excels across all tasks.

\begin{table*}[ht]
\centering
\resizebox{0.95\textwidth}{!}{
\begin{tabular}{c|ccccccccc}
\toprule
\multirow{2}{*}{\textsc{Model}} & \multicolumn{2}{c}{\textsc{\flores}} & \multicolumn{2}{c}{\textsc{\muse}} & \multicolumn{2}{c}{\textsc{\mlqa}} & \multirow{2}{*}{\textsc{\xquad}} & \multirow{2}{*}{\textsc{\sib}} & \multirow{2}{*}{\textsc{AVG}} \\
& \textsc{en-xx} & \textsc{xx-en} & \textsc{en-xx}
& \textsc{xx-en} & \textsc{en-xx} & \textsc{xx-en} & & & \\
\midrule
BLOOM-7B1 & 69.1 & 72.4 & 43.1 & 53.7 & 36.4 & 42.7 & 47.2 & 58.1 & 52.8 \\
+ Parallel Data & 70.3 & 72.3 & 45.5 & 53.9 & 38.0 & 43.9 & 48.3 & 59.5 & 54.0 \\
$\Delta$ & {\color{posred}01.2} & {\color{neggreen}-00.1} & {\color{posred}02.4} & {\color{posred}00.2} & {\color{posred}01.6} & {\color{posred}01.2} & {\color{posred}01.0} & {\color{posred}01.4} & {\color{posred}01.2} \\
\midrule
BLOOM-3B & 64.0 & 68.9 & 39.7 & 50.9 & 29.4 & 26.2 & 32.7 & 54.5 & 45.8 \\
+ Parallel Data & 65.0 & 69.1 & 41.4 & 51.6 & 30.9 & 26.7 & 34.5 & 56.9 & 47.0 \\
$\Delta$ & {\color{posred}01.0} & {\color{posred}00.2} & {\color{posred}01.8} & {\color{posred}00.7} & {\color{posred}01.5} & {\color{posred}00.5} & {\color{posred}01.8} & {\color{posred}02.4} & {\color{posred}01.2} \\
\midrule
BLOOM-1B7 & 59.0 & 65.8 & 37.2 & 48.5 & 20.0 & 22.2 & 24.8 & 53.0 & 41.3 \\
+ Parallel Data & 61.1 & 65.7 & 38.9 & 48.0 & 20.8 & 20.9 & 24.4 & 53.0 & 41.6 \\
$\Delta$ & {\color{posred}02.0} & {\color{neggreen}-00.1} & {\color{posred}01.6} & {\color{neggreen}-00.6} & {\color{posred}00.8} & {\color{neggreen}-01.3} & {\color{neggreen}-00.3} & {\color{posred}00.0} & {\color{posred}00.3} \\
\bottomrule
\end{tabular}
}
\caption{Effect of parallel corpora on BLOOM models of different sizes across various tasks. `+ Parallel Data' indicates instruction tuning of the given mLLM with the \mt objective, using 10K parallel corpora obtained with the best filtering strategy (ID 5) as shown in Table~\ref{tab:quality}.}
\label{tab:model_size}
\end{table*}

\section{Model Size}
\label{sec:model_size}

We explore the impact of parallel corpora on various sizes of BLOOM models, detailed in Table~\ref{tab:model_size}.

\textbf{Smaller models exhibit more pronounced improvements in \flores.} Notably, BLOOM-1B7 demonstrates larger improvements compared to its larger counterparts in the \flores task, where the prompt is the same as the one used during instruction tuning with the \mt objective. This is attributed to the smaller models' less developed in-context learning capabilities before instruction tuning, allowing for more substantial improvements when supplemented with parallel corpora.

\textbf{Larger models excel in diverse tasks.} Conversely, larger models generally demonstrate greater enhancements in tasks beyond machine translation. Both BLOOM-7B1 and BLOOM-3B exhibit a 1.2\% improvement compared to their original mLLMs, while BLOOM-1B7 shows a slight 0.3\% improvement. Specifically, BLOOM-7B1 and BLOOM-3B display notable improvements in tasks except for \flores, while BLOOM-1B7 achieves comparable or even worse results.

These findings demonstrate that when leveraging parallel corpora to enhance mLLMs, larger models not only exhibit improvements in direct tasks, such as machine translation, but also demonstrate a more substantial overall enhancement across a variety of tasks. In contrast, smaller models primarily show benefits in direct tasks. This difference can be attributed to the superior cross-task transferability of larger mLLMs, where insights gained from parallel corpora in one task contribute to improved performance in others.
 
\section{Conclusion}

This paper investigates the impact of four critical factors -- data quality, data quantity, objectives, and mLLM sizes -- on leveraging parallel corpora to enhance mLLMs across diverse languages and tasks. Our findings underscore the crucial importance of filtering out noisy translations to procure high-quality training data for improving mLLMs. Surprisingly, even a relatively modest dataset of 10K samples can yield promising results. Furthermore, our analysis shows that employing the machine translation objective leads to optimal outcomes. Importantly, larger models exhibit a greater capacity to benefit from parallel corpora, achieving more substantial improvements. This study provides a comprehensive recipe for effectively leveraging parallel corpora to enhance mLLMs. These insights significantly contribute to advancing the understanding and optimization of mLLMs across different languages and tasks.

\section*{Limitations}

Due to limited computational resources, we opted not to explore full-parameter training for leveraging parallel corpora. Instead, we focused on LoRA, drawing on insights from previous studies. Additionally, our investigation is restricted to the BLOOM series, and we did not extend our analysis to other mLLMs. Furthermore, we did not also explore mLLMs larger than 7B1.

\section*{Acknowledgements}
This work was funded by DFG (SCHU 2246/14-1),
the European Research Council (DECOLLAGE, ERC-2022-CoG \#101088763),  EU's Horizon Europe Research and Innovation Actions (UTTER, contract 101070631), by the Portuguese Recovery and Resilience Plan through project C645008882-00000055 (Center for Responsible AI),  by the DAAD programme Konrad Zuse Schools of Excellence in Artificial Intelligence, sponsored by the Federal Ministry of Education and Research, 
and by FCT/MECI through national funds and when applicable co-funded EU funds under UID/50008: Instituto de Telecomunicações. 

\bibliography{anthology,custom}

\clearpage
\appendix
\input{appendix_xglm}
\input{appendix_lora}

\input{appendix_prompt}

\end{document}

%% file: appendix_xglm.tex
\section{Additional Experiments on XGLM}
\label{sec:xglm}

We explore the effect of parallel corpora quality on XGLM-7.5B, and the results are shown in Table~\ref{tab:xglm}. As shown, experiments on XGLM-7.5B exhibits consistent findings as those on BLOOM-7B1.

\begin{table}[ht]
\centering
\begin{tabular}{c|ccc|c}
\toprule
\multirow{1}{*}{\textsc{ID}} & \multicolumn{3}{c|}{\multirow{1}{*}{\textsc{Model}}} & Accuracy \\
\midrule
0 & \multicolumn{3}{c|}{XGLM-7.5B} & 52.8 \\
\midrule
& $\tau_c$ & $\tau_l$ & LID & \\
\midrule
1 & 0 & 0 & \cmark & 54.6 \\
2 & 0.5 & 0 & \cmark & 55.0 \\
3 & 0.75 & 5 & \cmark & 54.4 \\
4 & 0.75 & 0 & \xmark & 55.0 \\
5 & 0.75 & 0 & \cmark & \textbf{55.5} \\
\bottomrule
\end{tabular}
\caption{Performance (\%) of XGLM-7.5B on SIB after instruction tuning with 
the machine translation objective using 10K parallel corpora with various quality filtering strategies. Parameters include $\tau_c$ for COMETWIKI score threshold, $\tau_l$ for sentence length threshold, and LID indicating the adoption of language identification filtering.}
\label{tab:xglm}
\end{table}

%% file: appendix_lora.tex
\section{Effect of LoRA Rank}
\label{sec:lora}

We conduct initial experiments on Arabic to explore the effect of setting the rank of LoRA, and the results are shown in Table~\ref{tab:lora}. As shown, setting the rank of LoRA as 16 lead to best performance. Therefore, we use LoRA rank as 16 across all the experiments.

\begin{table}[]
    \centering
    \begin{tabular}{c|c}
        \toprule
        Rank & Accuracy \\
        \midrule
        8 & 59.0 \\
        16 & \textbf{62.0} \\
        32 & 60.0 \\
        128 & 61.0 \\
        256 & 60.5 \\
        512 & 59.5 \\
        \bottomrule
    \end{tabular}
    \caption{Effect of LoRA rank: Accuracy on SIB using English-Arabic parallel data to improve BLOOM-7B1.}
    \label{tab:lora}
\end{table}

%% file: appendix_prompt.tex
\section{Prompt}
\label{sec:prompt}

The prompts of \flores, \muse, \mlqa, \xquad, and \sib are shown as follows: %

\paragraph{\flores/\muse}
\begin{quote}
    \texttt{Translate the following text from [SOURCE\_LANG] to [TARGET\_LANG].\textbackslash nText: [SOURCE\_TEXT]\textbackslash nTranslation: [TARGET\_TEXT]}
\end{quote}

\paragraph{\mlqa/\xquad}

\begin{quote}
\texttt{[Passage] \textbackslash nQ: [Question]\textbackslash nA: [Answer]}
\end{quote}

\paragraph{\sib}

\begin{quote}
\texttt{The topic of the news [Passage] is [Label]}
\end{quote}